\pdfoutput=1

\documentclass[11pt]{article}

\usepackage[preprint]{acl}

\usepackage{times}
\usepackage{latexsym}
\usepackage{seqsplit}

\usepackage{tabularx}
\usepackage{booktabs}
\usepackage{array}

\usepackage[T1]{fontenc}

\usepackage[utf8]{inputenc}

\usepackage{microtype}

\usepackage{inconsolata}

\usepackage{graphicx}

\usepackage{amsmath}
\usepackage{amssymb}
\usepackage{mathtools}
\usepackage{amsthm}

\usepackage{booktabs}
\usepackage{cleveref}
\usepackage{graphicx}
\usepackage{multirow}
\usepackage{tablefootnote}
\usepackage{caption}
\usepackage{subcaption}
\usepackage{url}
\usepackage{makecell}
\usepackage{color,colortbl}
\usepackage{enumitem}
\usepackage{hyperref}
\usepackage{listings}
\usepackage{xspace}
\usepackage{balance}

\newcommand\blfootnote[1]{%
  \begingroup
  \renewcommand\thefootnote{}\footnote{#1}%
  \addtocounter{footnote}{-1}%
  \endgroup
}

\newcommand{\modelname}{\texttt{MirrorAPI}\xspace}
\newcommand{\modelnamebench}{\texttt{MirrorAPI-Bench}\xspace}

\title{StableToolBench-\texttt{MirrorAPI}: \\
Modeling Tool Environments as Mirrors of 7,000+ Real-World APIs
}

\renewcommand{\thefootnote}{\fnsymbol{footnote}}

\author{
Zhicheng Guo$\textsuperscript{\rm $1,2 *$}$\footnotemark[2],
Sijie Cheng$\textsuperscript{\rm $1,2,3*$}$, 
Yuchen Niu,
Hao Wang$\textsuperscript{\rm $4$}$,\\
\textbf{Sicheng Zhou}$^{5}$,
\textbf{Wenbing Huang}$^{6}$,
\textbf{Yang Liu}$^{1,2}$
\\
 $\textsuperscript{\rm $1$}$Dept. of Comp. Sci. \& Tech., Institute for AI, Tsinghua University, Beijing, China\\
 $\textsuperscript{\rm $2$}$Institute for AI Industry Research (AIR), Tsinghua University, Beijing, China \\
 $\textsuperscript{\rm $3$}$RayNeo
 $\textsuperscript{\rm $4$}$Google
 $\textsuperscript{\rm $5$}$The University of Toronto, Canada\\
  $\textsuperscript{\rm $6$}$Gaoling School of Artificial Intelligence, Renmin University of China
 \\
 \{\texttt{guo-zc21}, \texttt{csj23}\}\texttt{@mails.tsinghua.edu.cn}
}

\begin{document}
\maketitle
\footnotetext[2]{Project Leader \textsuperscript{*}Equal contribution}
\begin{abstract}
The rapid advancement of large language models (LLMs) has spurred significant interest in tool learning, where LLMs are augmented with external tools to tackle complex tasks. However, existing tool environments face challenges in balancing stability, scalability, and realness, particularly for benchmarking purposes. To address this problem, we propose \modelname, a novel framework that trains specialized LLMs to accurately simulate real API responses, effectively acting as ``mirrors'' to tool environments. Using a comprehensive dataset of request-response pairs from 7,000+ APIs, we employ supervised fine-tuning and chain-of-thought reasoning to enhance simulation fidelity. \modelname achieves superior accuracy and stability compared to state-of-the-art methods, as demonstrated by its performance on the newly constructed \modelnamebench and its integration into StableToolBench.

\blfootnote{GitHub: \href{https://github.com/THUNLP-MT/StableToolBench}{THUNLP-MT/StableToolBench}}
\blfootnote{Dataset and Models: \url{https://huggingface.co/stabletoolbench}}

\end{abstract}
\renewcommand{\thefootnote}{\arabic{footnote}}
\section{Introduction}

Recently, the remarkable reasoning capabilities exhibited by large language models (LLMs)~\citep{qwen2025qwen25technicalreport,deepseekai2025deepseekr1incentivizingreasoningcapability,llama3modelcard,geminiteam2024geminifamilyhighlycapable} have catalyzed growing interest in tool learning~\cite{gpt4tools,ye2024tooleyes,seal-tools}. Analogous to the way humans leverage tools in the physical world to enhance their cognitive and physical capacities, LLMs augmented with external tools~\citep{Qu_2025,qin2024toollearningfoundationmodels} demonstrate unprecedented potential in addressing diverse and complex challenges such as task solving~\cite{qin2023tool,huang2024understandingplanningllmagents}, web search~\citep{komeili-etal-2022-internet,gou2024critic}, and multimodal scenarios~\citep{ma-etal-2024-sciagent,wang2024mllmtoolmultimodallargelanguage}.

In a typical tool-using scenario, three main components are involved: (1) a user who provides instructions, (2) a tool-using model that performs actions, and (3) an environment containing tools that respond to those actions.
The environment plays a critical role, as it directly influences the quantity, quality, and stability of available tools.
Most environments~\citep{tang2023toolalpaca, chen-etal-2024-eval,basu-etal-2024-api,farn2023tooltalk} use the real-world, publicly-available tools~\footnote{In this work, we use the terms APIs and tools interchangeably.}, such as RapidAPI~\footnote{\url{https://rapidapi.com/}}.
Others use manually selected or created APIs~\cite{li2023apibank, xu2023tool}. They often prioritise stable APIs like Google services or Python libraries. Recent studies further explore prompting LLMs to generate APIs and simulate their behaviours according to documentation.~\citep{utraTool, seal-tools}. 


\begin{figure}
    \centering
    \includegraphics[width=\linewidth]{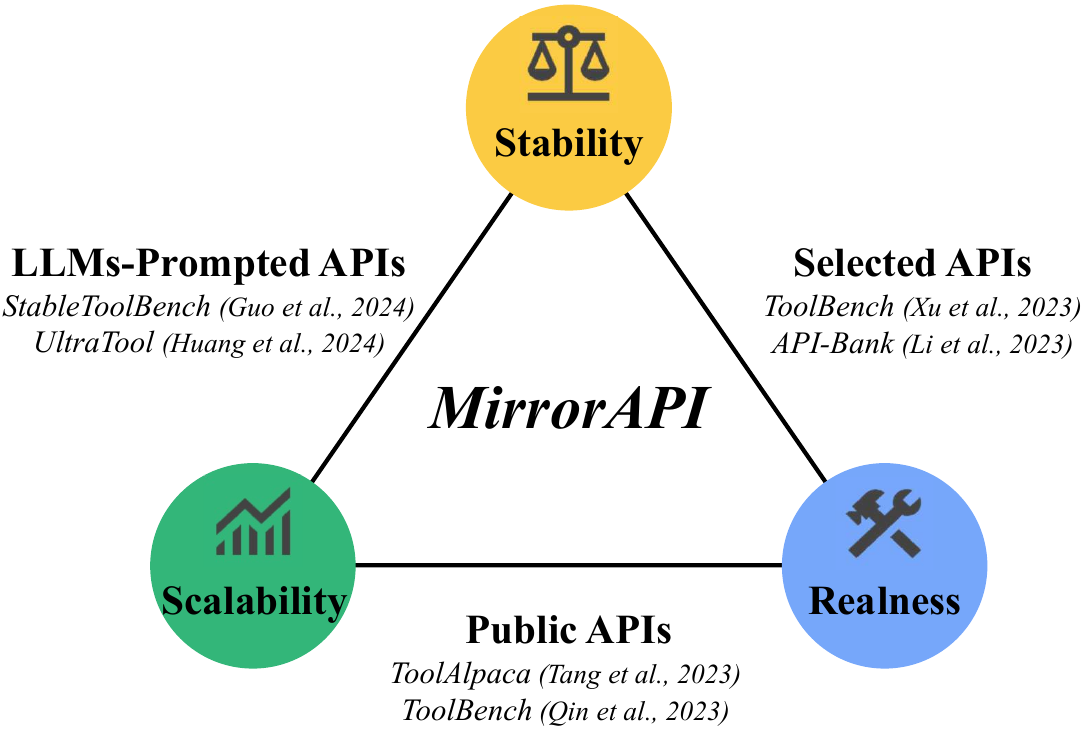}
    \caption{While existing API environments prioritize one or two of stability, scalability, or realness, \modelname effectively harmonizes all three aspects.}
    \label{fig:triangle}
\end{figure}

However, existing environments struggle to balance stability, scalability and realness.
Environments built on large-scale public APIs often exhibit instability, leading to inconsistent results over time~\citep{guo-etal-2024-stabletoolbench}.
This instability stems from developer updates, underlying behavioural modifications, and network connectivity fluctuations.
On the other hand, environments relying on manually selected or created APIs lack scalability due to limited labour resources.
While LLMs-simulated APIs offer greater stability compared to real-world APIs, there is a significant gap between simulated behaviours and actual API response, as demonstrated in our experiments in \Cref{sec:compare_real_sim}.



\begin{figure*}
    \centering
    \includegraphics[width=0.85\linewidth]{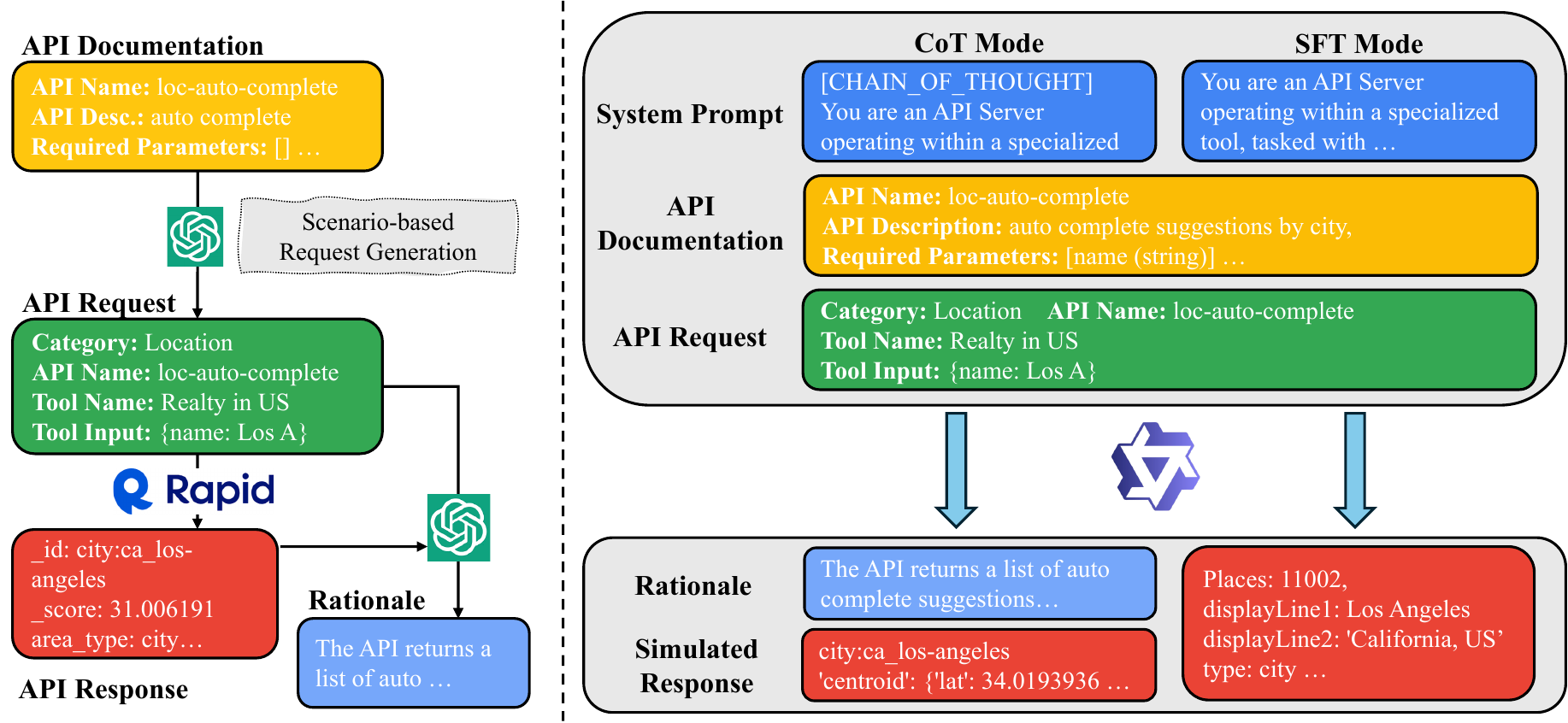}
    \caption{The phases of data construction (left) and the training of \modelname (right).}
    \label{fig:method}
\end{figure*}

To balance stability, scalability and realness of tool-learning environments, we propose \modelname, a novel framework that trains specialized LLMs to act as ``mirrors'' to tool environments by accurately replicating real API responses.  
A tool can be conceptualized as a wrapper around its underlying system, abstracted through manually crafted documentation.
Given that real documentation data is both informative and practical to collect, it is natural to simulate tool behaviour based on documentation and user requests.
However, the intricate logic of complex underlying systems cannot be fully captured in documentation alone.
Therefore, it is crucial to align simulated behaviour not only with documentation but also with the practical workings of real-world APIs. 


To train the specialized LLMs, named \modelname, we first collect a dataset of real request-response pairs spanning 49 categories and 7,000+ APIs with documentation from RapidAPI for supervised fine-tuning (SFT).
To further capture the implicit factors in underlying real API systems, we incorporate reasoned explanations of API mechanisms into the training data. 
Specifically, given request-response pairs, we use OpenAI \texttt{o1-preview} to generate chain-of-thought (CoT) rationales that explain the underlying working mechanisms of APIs~\cite{yang2024reactmeetsactrelanguage}.  
The \modelname is trained in both SFT and CoT modes, using special prompt tokens to distinguish~\citep{wang2023openchat}.
During inference, the model defaults to SFT mode for consistent and better performance.

To evaluate the model, we construct \modelnamebench by defining an in-distribution (ID) and an out-of-distribution (OOD) sets based on real request-response pairs with APIs seen and unseen during training.
Experiments demonstrate that \modelname excels in simulating real APIs, outperforming state-of-the-art LLM prompting methods in both documentation and instruction following capabilities.
Additionally, \modelname achieves the highest similarity to real responses. 
Second, we integrate \modelname as the tool environment into StableToolBench~\cite{guo-etal-2024-stabletoolbench}. 
Results show that \modelname not only maintains full environment stability but also produces outcomes comparable to real environments. 
Beyond benchmarking, \modelname can enhance tool-using models by providing step-wise feedback or expanding the size and diversity of training data, offering a promising direction for future research.


\section{Methods}

\subsection{Request-Response Pairs Collection}
To train specialized LLMs for simulating real APIs, we collect training data by calling real RapidAPI servers using LLMs-generated requests.
The process consists of three stages: (1) collecting real request-response pairs, (2) filtering request-response pairs, and (3) synthesizing request-response pairs.

\subsubsection{Collecting Real Request-Response Pairs}
We first construct a large collection of APIs and uniformly call them to gather data. 
The data collection process is illustrated in \Cref{fig:method}.
Following ToolBench~\citep{qin2023tool}, we crawl the latest tools and their documentation from RapidAPI.
Moreover, we perform dummy calls to verify the availability of these APIs. 
Then, we exploit LLMs to generate API requests for the available APIs. 
When directly prompted to write requests based on API documentation, LLMs tend to produce simple calls with limited diversity. To enhance data complexity and diversity, we employ a two-stage scenario-based approach as shown in \Cref{app:request_writing}.
First, we prompt LLMs to generate realistic usage scenarios for a given API based on its documentation, setting the temperature to 1.0 to maximize diversity.
Next, we prompt LLMs to write API requests that address the tasks described in these scenarios, setting the temperature to 0.1 to ensure stability and precision.
Finally, we generate 200 calls for each available API using the predefined method and record the responses. Each request was attempted up to three times in case of failures.

\subsubsection{Filtering Request-Response Pairs} \label{sec:postprocess}
First, real API responses often include unsuccessful calls due to various reasons, such as incorrect requests generated by LLMs, discrepancies between locally crawled API documentation and real-time API behaviour, and temporary API unavailability. To mitigate these issues, we filter out unsuccessful calls using the same rules outlined in StableToolBench as discussed in \Cref{app:filter_rule}.
Second, to ensure the trained simulator can return failure messages for requests that violate documentation requirements, we retain around 10,000 unsuccessful calls caused by parameter-related issues, such as missing required parameters or incorrect parameter values.
Finally, due to the large number of poorly written documentation, some real API responses may deviate from their documentation, which can harm model performance.
To mitigate this, we elicit \texttt{gpt-4o} with the prompt as shown in Table~\ref{tab:prompt_check_document_follow}. This effectively filters out request-response pairs that do not align with their documentation.


\subsubsection{Synthesizing Request-Response Pairs}
To further enlarge and balance the dataset, we use \texttt{gpt-4o} to generate synthetic request-response pairs based on existing examples. This serves two purposes: (1) to augment the dataset size, ensuring sufficient data for robust model training, and (2) to mitigate the scarcity of data samples for long-tail APIs.
For each generation, the prompt includes a list of 10 real API examples, comprising API documentation, LLM-generated requests, and real responses. The model is encouraged to produce diverse request-response pairs by varying temperature and seed values. 
An exact match function ensures that each generated pair is unique and does not already exist in the dataset.
After generation, we conduct a quality check and filter out pairs with identical requests but differing responses, or responses inconsistent with the documentation.

\subsubsection{Statistics and \modelnamebench}
In total, we collect 95,872 API request-response pairs, spanning 7,500 APIs across all 49 categories in RapidAPI.
Then, we partition the dataset into training and testing sets, named \modelnamebench. 
Specifically, the OOD testing set comprises 200 successful and 100 unsuccessful pairs from 49 APIs not included in the training data.
Additionally, we construct three ID test sets, each consisting of 100 successful request-response pairs drawn from the training APIs, representing the high-resource, medium-resource, and low-resource groups.
These sets correspond to the top third, middle third, and bottom third of APIs, respectively, as illustrated in \Cref{fig:tool_category_count}.
The detailed statistics of the datasets are summarized in \Cref{tab:dataset_statistics}.

\begin{table}[t]
    \centering
    \small
    \begin{tabular}{lccc}
        \toprule
        \textbf{Dataset} & \textbf{Samples} & \textbf{APIs} & \textbf{Categories} \\
        \midrule
        Training & 95,272 & 7,437 & 49 \\
        OOD Succ & 200 & 38 & 21 \\
        OOD Fail & 100 & 22 & 14 \\
        ID High & 100 & 96 & 14 \\
        ID Medium & 100 & 97 & 15 \\
        ID Low & 100 & 81 & 14 \\
        \bottomrule
    \end{tabular}
    \caption{Dataset Statistics}
    \label{tab:dataset_statistics}
\end{table}


\begin{table*}[ht!]
    \centering
    \small
    \begin{tabular}{lcccccc}
        \toprule
        \textbf{Models} & \textbf{OOD Succ} & \textbf{OOD Fail} & \textbf{ID High} & \textbf{ID Medium} & \textbf{ID Low} \\
        \midrule

        GPT-4o & 5.93 & 2.06 & 4.58 & 4.76 & 3.92 \\
        GPT-4o mini & 6.10 & 1.95 & 5.17 & 4.09 & 3.57 \\
        GPT-4o CoT & 4.45 & 3.72 & 4.06 & 4.26 & 4.20 \\
        o1-preview & \textbf{7.67} & 1.81 & 6.11 & 5.47 & 4.11 \\
        Llama 3.1 8B & 1.27 & 1.86 & 1.74 & 1.61 & 2.28 \\
        Qwen2.5 7B & 2.99 & 1.71 & 3.33 & 2.89 & 2.27 \\
        Deepseek-R1-Distill-Qwen-32B & 5.97 & 1.82 & 4.61 & 4.33 & 3.38 \\
        Deepseek-R1-Distill-Qwen-7B & 3.03 & 1.83 & 2.42 & 3.12 & 2.56 \\
        \midrule
        \modelname CoT & 6.51 & \textbf{8.64} & \textbf{8.91} & \textbf{8.79} & \textbf{9.01} \\
        \modelname SFT & \underline{6.86} & \underline{8.28} & \underline{8.76} & \underline{8.63} & \underline{8.74} \\
        \bottomrule
    \end{tabular}
    \caption{Performance of observation following on \modelnamebench. OOD Succ and OOD Fail stand for the test sets of OOD successful and unsuccessful calls respectively. The ID High/Medium/Low stands for the test sets of ID calls with a large/medium/low number of training samples. Tables below use the same notation.}
    \label{tab:perf_observation_following}
\end{table*}

\subsection{Chain-of-Thought Data Collection}
Simply training models to follow API documentation does not fully close the gap between the simulated and real APIs as the model is ignorant of the working mechanism behind the real APIs. Therefore, we propose to improve the simulation model by augmenting reasoning API mechanism. The data collection process is illustrated in \Cref{fig:method}.

Inspired by the ActRe~\cite{yang2024reactmeetsactrelanguage}, we fully utilize the real responses in the data. Specifically, for a request-response pair $(R,O)$, where $R$ is a request to an API and $O$ is the response from the API, we use a general reasoning LLM (OpenAI O1) to reason the implicit working mechanism, which serves as a rationale for simulation. The prompt is shown in Table~\ref{tab:prompt_rationale_generation}, i.e.,
\begin{equation*}
    R_{o1}=LLM_{o1}(R,O,Doc)
\end{equation*}
where $Doc$ denotes the API documentation.

We randomly sample 43,597 successful call-response pairs and 2,000 unsuccessful pairs from the training set to generate rationales. To ensure the quality of the generated rationales and avoid excessively verbose explanations, we apply a length filter, restricting the maximum token count of the final outputs to 2,560. As a result, we obtain 42,465 successful and 1,975 unsuccessful request-response pairs, each annotated with a generated rationale. Note that all CoT data is constructed based on the SFT training data, meaning no request-response data is created here. 

\subsection{Model Training}
We train the environment model in the both SFT and CoT modes simultaneously. Namely, we mix 95,272 SFT samples and 44,440 CoT samples in the final training data. 
This combination ensures that the \modelname benefits from the extensive SFT data while also leveraging the performance-boosting advantages of CoT. Illustration of the two modes can be seen in \Cref{fig:method}.

To distinguish the two modes during training and testing, following OpenChat~\citep{wang2023openchat}, we introduce a special tag \texttt{[chain-of-thought]} at the beginning of the CoT mode system prompt. This approach allows the simulator to operate in two distinct modes by switching between system prompts. 
The prompts for the SFT and CoT cases are shown in Table~\ref{tab:prompt_system_sft} and \ref{tab:prompt_system_cot}, respectively.

During training, we finetune a \texttt{\seqsplit{Qwen2.5-7B-Intruct}}~\citep{qwen2025qwen25technicalreport} model to generate API responses in the SFT mode, given the user request and API documentation. In the CoT mode, the model is trained to generate both the rationale and the simulation output. Formally, these two modes can be expressed as:
\[
\begin{aligned}
S &= LLM_{simu}(R, Doc | P_{SFT}) &&\text{(\textit{SFT mode})}\\
[R_{o1}, S] &= LLM_{simu}(R, Doc | P_{CoT}) &&\text{(\textit{CoT mode})}
\end{aligned}
\]
where $S, R, Doc$ and $LLM_{simu}$ represent the simulation output, the user request, the API documentation and the LLM to train, respectively. $P_{SFT}$ and $P_{CoT}$ are the system prompts used in the two modes. $R_{o1}$ is the rationale generated by \texttt{o1-preview}.

At testing time, the model generates simulations based on user requests and API documentation in the SFT mode. In the CoT mode, it additionally produces rationales before delivering final outputs. Mathematically, this can be represented as:
\[
\begin{aligned}
S &= LLM_{simu}(R, Doc | P_{SFT}) &&\text{(\textit{SFT mode})}\\
[R_s, S] &= LLM_{simu}(R, Doc | P_{CoT}) &&\text{(\textit{CoT mode})}
\end{aligned}
\]
where $S, R, Doc$ and $LLM_{simu}$ represent the simulation output, the user request, the API documentation and the trained model respectively. $P_{SFT}$ and $P_{CoT}$ are the system prompts used in the two modes. $R_s$ is the rationale the model generates before giving a simulation output. Note that during inference, we set the SFT mode as default.

\begin{table*}[t!]
    \centering
    \small
    \begin{tabular}{lcccccccccc}
        \toprule
        \multirow{2}{*}{\textbf{Models}} & 
        \multicolumn{2}{c}{\textbf{OOD Succ}} & 
        \multicolumn{2}{c}{\textbf{OOD Fail}} & 
        \multicolumn{2}{c}{\textbf{ID High}} & 
        \multicolumn{2}{c}{\textbf{ID Medium}} & 
        \multicolumn{2}{c}{\textbf{ID Low}} \\
        \cmidrule(lr){2-3} \cmidrule(lr){4-5} \cmidrule(lr){6-7} \cmidrule(lr){8-9} \cmidrule(lr){10-11}
        & BLEU & Cosine & BLEU & Cosine & BLEU & Cosine & BLEU & Cosine & BLEU & Cosine \\
        \midrule
        GPT-4o & 13.4 & 63.1 & 10.7 & 36.0 & 10.0 & 55.8 & 12.4 & 53.4 & 14.8 & 52.8  \\
        GPT-4o mini & 14.7 & 62.3 & 11.3 & 36.6 & 10.3 & 56.2 & 12.5 & 53.1 & 14.5 & 53.6 \\
        GPT-4o CoT & 13.4 & 60.3 & 13.6 & 42.8 & 11.7 & 55.0 & 14.2 & 52.8 & 16.5 & 54.5 \\
        o1-preview & 19.2 & 65.6 & 8.4 & 37.2 & 12.1 & 57.5 & 14.6 & 55.7 & 15.9 & 54.3 \\
        Llama 3.1 8B & 5.3 & 49.6 & 11.2 & 43.7 & 6.3 & 47.2 & 9.6 & 46.3 & 9.9 & 47.8 \\
        Qwen 2.5 7B Instruct & 7.5 & 53.5 & 9.2 & 36.3 & 7.7 & 50.6 & 11.3 & 48.4 & 7.7 & 47.3 \\
        Deepseek-32B & 12.6 & 63.6 & 9.7 & 38.1 & 9.2 & 55.5 & 13.4 & 53.8 & 15.9 & 53.9 \\
        Deepseek-7B & 12.3 & 60.0 & 12.0 & 38.0 & 10.8 & 52.3 & 12.6 & 51.9 & 12.8 & 51.8 \\
        \midrule
        \modelname CoT & \underline{31.6} & \underline{66.7} & \underline{84.9} & \underline{90.7} & \underline{73.6} & \underline{87.6} & \underline{77.5} & \underline{88.0} & \underline{84.7} & \underline{93.4} \\
        \modelname SFT & \textbf{35.6} & \textbf{69.9} & \textbf{89.9} & \textbf{94.1} & \textbf{74.2} & \textbf{88.7} & \textbf{80.0} & \textbf{89.8} & \textbf{86.3} & \textbf{94.2} \\
        \bottomrule
    \end{tabular}
    \caption{Comparison of BLEU scores and LLM cosine similarity between real and simulated responses across different groups. BLEU and Cosine stand for the BLEU metric and LLM consine similarity metrics respectively. Deepseek-32B and Deepseek-7B are Deepseek-R1-Distill-Qwen-32B and Deepseek-R1-Distill-Qwen-7B.}
    \label{tab:combined_scores_bleu_cosine}
\end{table*}


\section{Evaluation on \modelnamebench}
\subsection{Evaluation Metrics}
\noindent \textbf{Following Documentation and Instructions.}
To assess the documentation and instruction following capability of the simulation model, we adopt an LLM-as-a-judge framework, following FastChat~\cite{zheng2023judging}. The prompt is shown in Table~\ref{tab:prompt_llm_judge}. Specifically, we employ \texttt{gpt-4o} as the evaluator model, judging whether the simulated responses align with the documentation and user requests, rating from 1 to 10.

\noindent \textbf{Similarity to Real Responses.}
In addition to evaluating the following capability, we directly compare the simulated responses with those real responses from real-world APIs. 
To do so, we adopt two metrics: BLEU~\cite{papineni-etal-2002-bleu} and LLM cosine similarity~\cite{bert-score}. 
Specifically, we calculate BLEU-4 scores from NLTK toolkit~\cite{bird2009natural}. 
For the LLM cosine similarity, we first encode the real API responses and simulated responses with OpenAI \texttt{text-embedding-3-small}, respectively. Then we calculate the cosine similarity score between the real and simulated responses pairwise. Mathematically,
\begin{equation*}
    S =  \frac{1}{N} \sum_{n=1}^N\cos(Enc(R_{real}), Enc(R_{simu}))
\end{equation*}
where $S,N, R_{real}, R_{simu}$ and $Enc$ are the similarity score, total number of samples in the test set, real responses, simulated responses, and the encoding LLMs, respectively.

\subsection{Baseline}
We evaluate several baselines on the simulation test data, including both general-purpose and reasoning-enhanced LLMs. 
These include \texttt{gpt-4o} and \texttt{gpt-4o-mini}~\citep{openai2024gpt4o}, which are API-based, general-purpose LLMs. Additionally, we consider \texttt{gpt-4o} CoT, a variant of \texttt{gpt-4o} integrated with chain-of-thought reasoning through the API mechanism.
We also include \texttt{o1-preview}~\citep{openai2024o1preview}, a general-purpose LLM known for its strong reasoning capabilities. 
For open-source alternatives, we evaluate \texttt{Llama-3.1-8B-instruct}~\citep{llama3modelcard} and \texttt{Qwen-2.5-7B-Intruct}~\citep{qwen2025qwen25technicalreport}, both of which are general-purpose LLMs. 
Finally, we assess the \texttt{Deepseek-R1-Distill-Qwen-7B} and \texttt{Deepseek-R1-Distill-Qwen-32B}~\citep{deepseekai2025deepseekr1incentivizingreasoningcapability}, which are open-source models with enhanced reasoning abilities.

\subsection{Results}\label{sec:results}

We run our model with both CoT and SFT prompts. We keep the temperature at 0 during generation for all models. 
The results for the documentation and instruction-following metrics are presented in \Cref{tab:perf_observation_following}. BLEU scores and LLMs Mean Cosine Similarity scores are detailed in \Cref{tab:combined_scores_bleu_cosine}.
As shown in the tables, our trained simulation model outperforms all baselines, except for \texttt{o1-preview}, on OOD successful tasks when evaluated using the documentation and instruction-following metrics, demonstrating the generalizability of our approach.
Moreover, our model surpasses all other baselines, including \texttt{o1-preview}, on OOD failing tasks and all ID sub-tasks across all metrics. This result underscores the effectiveness of training on API calls in enabling accurate simulation of those APIs. Interestingly, \texttt{o1-preview} outperforms all prompting methods on the OOD successful tasks, showing the potential of reasoning for simulation.
Additionally, our trained SFT model outperforms its CoT counterparts on OOD successful tasks. However, this does not diminish the importance of reasoning-based training. Further insights and supporting evidence are explored in \Cref{sec:ablation}.

\section{Environment Simulation for StableToolBench}
An important application of the simulation model is its role as an environment model in a tool learning benchmark. 
To improve reliability and stability of simulation on StableToolBench, we further finetune \modelname on the cache data, which is \modelname-Cache.
To evaluate \modelname and \modelname-Cache on StableToolBench, we compare the real environment with our \modelname in the StableToolBench setting, with the Solvable Pass Rate (SoPR) and our proposed Final Answer Completeness (FAC) scores. Additionally, we present the performance of several baseline models when \modelname is used as the environment model, demonstrating its reliability.

\begin{figure}[t!]
    \centering
    \begin{subfigure}{\linewidth}
        \centering
        \includegraphics[width=\linewidth]{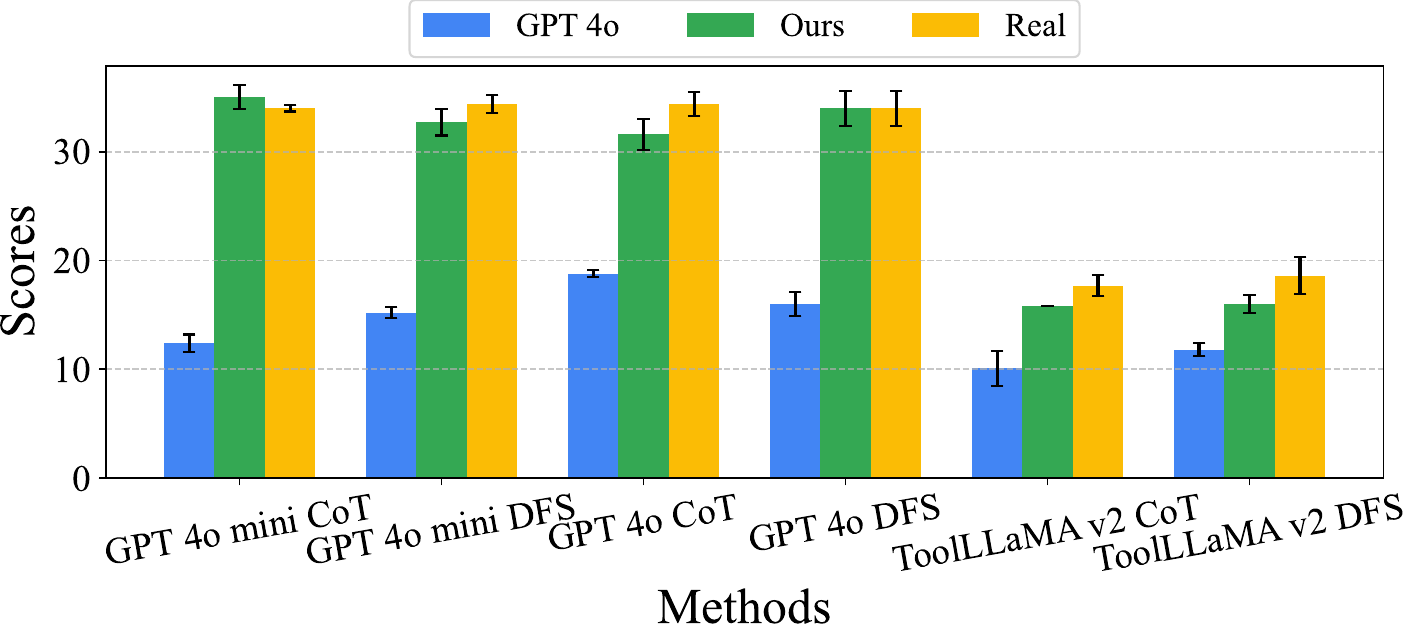}  
        \caption{Solvable pass rate scores. We run all models once, evaluate three times and take the average results. Error bars are the standard deviation during evaluation.}
        \label{fig:compare_pass_rate}
    \end{subfigure}
    \begin{subfigure}{\linewidth}
        \centering
        \includegraphics[width=\linewidth]{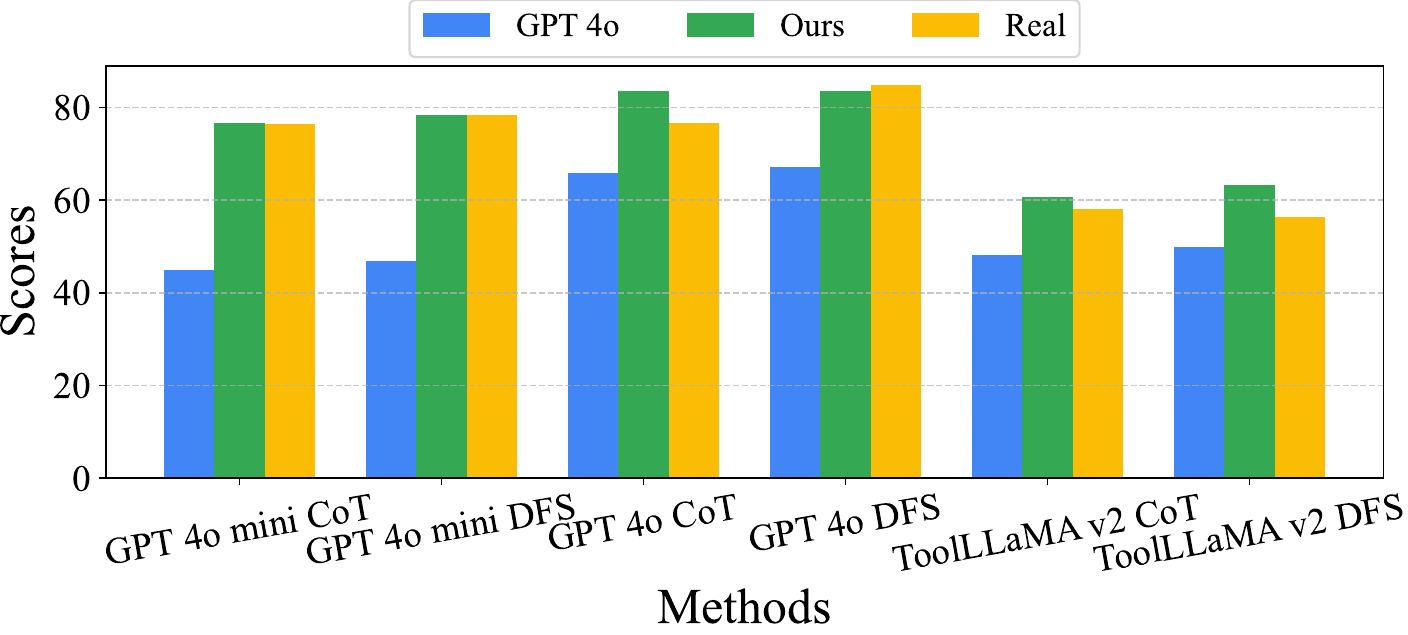}
        \caption{Final Answer Completeness scores.}
        \label{fig:compare_evaluator_model}
    \end{subfigure}
    \caption{Comparison between the real environment, GPT 4o, and \modelname simulated environments.}
    \label{fig:compare_all}
\end{figure}

\subsection{Training \modelname-Cache}\label{sec:method_cache}
To better improve the performance of the model on benchmarks, we fine-tune the model using real API data accumulated from the StableToolBench. This dataset provides a rich collection of real-world API usage data, which is essential for evaluating the model's performance in practical scenarios. By leveraging this dataset, we aim to ensure the model is better aligned with actual user needs and behaviours, thereby improving its ability to generalize to real-world tasks. The dataset used for fine-tuning comprises 110,700 samples, in which $200$ of the samples are used as the test set. 
\begin{table*}[!ht]
    \centering
    \small
    \begin{tabular}{lccccccc}
        \toprule
        \textbf{Method} & \textbf{I1 Inst} & \textbf{I1 Cat} & \textbf{I1 Tool} & \textbf{I2 Cat} & \textbf{I2 Inst} & \textbf{I3 Inst} & \textbf{Average} \\
        \midrule
        ToolLLaMA v2 CoT & 28.0{\tiny $\pm{1.9}$} & 30.5{\tiny $\pm{0.8}$} & 21.5{\tiny $\pm{0.9}$} & 19.9{\tiny $\pm{1.0}$} & 22.3{\tiny $\pm{0.4}$} & 19.1{\tiny $\pm{0.8}$} & 22.8{\tiny $\pm{0.8}$} \\
    ToolLLaMA v2 DFS & 28.4{\tiny $\pm{0.9}$} & 32.5{\tiny $\pm{0.8}$} & 22.2{\tiny $\pm{1.0}$} & 22.8{\tiny $\pm{1.5}$} & 19.2{\tiny $\pm{1.6}$} & 18.6{\tiny $\pm{1.5}$} & 22.9{\tiny $\pm{1.4}$} \\
    GPT 4o mini CoT & 27.8{\tiny $\pm{1.4}$} & 34.9{\tiny $\pm{0.3}$} & 34.2{\tiny $\pm{0.5}$} & 24.5{\tiny $\pm{1.0}$} & 22.3{\tiny $\pm{2.7}$} & 20.8{\tiny $\pm{1.5}$} & 25.9{\tiny $\pm{1.7}$} \\
    GPT 4o mini DFS & 26.8{\tiny $\pm{1.4}$} & 36.4{\tiny $\pm{1.6}$} & 33.1{\tiny $\pm{1.1}$} & 25.8{\tiny $\pm{1.7}$} & 25.8{\tiny $\pm{2.7}$} & 20.2{\tiny $\pm{0.8}$} & 26.4{\tiny $\pm{1.6}$} \\
    GPT 4o CoT & \textbf{33.3}{\tiny $\pm{2.0}$} & 35.1{\tiny $\pm{0.6}$} & 33.6{\tiny $\pm{0.8}$} & 32.5{\tiny $\pm{1.7}$} & \textbf{29.6}{\tiny $\pm{1.6}$} & \textbf{27.9}{\tiny $\pm{3.5}$} & \textbf{32.0}{\tiny $\pm{2.2}$} \\
    GPT 4o DFS & 32.7{\tiny $\pm{1.9}$} & \textbf{42.3}{\tiny $\pm{1.3}$} & \textbf{34.6}{\tiny $\pm{1.3}$} & \textbf{32.8}{\tiny $\pm{1.5}$} & 28.3{\tiny $\pm{1.3}$} & 23.0{\tiny $\pm{1.3}$} & 30.9{\tiny $\pm{1.7}$} \\
    \bottomrule
    \end{tabular}
    \caption{Solvable pass rate scores.  ``Inst'' and ``Cat'' stand for the Instruction and Category subsets. Tables below use the same abbreviation. }
    \label{tab:main_sopr}
\end{table*}
\begin{table*}[!ht]
    \centering
    \small
    \begin{tabular}{lccccccc}
        \toprule
        \textbf{Method} & \textbf{I1 Inst} & \textbf{I1 Cat} & \textbf{I1 Tool} & \textbf{I2 Cat} & \textbf{I2 Inst} & \textbf{I3 Inst} & \textbf{Average} \\
        \midrule
        ToolLLaMA v2 CoT & 45.4 & 38.6 & 34.2 & 40.3 & 37.7 & 31.1 & 37.9 \\
    ToolLLaMA v2 DFS & \textbf{47.9} & 40.5 & 31.0 & 40.3 & 34.0 & 31.1 & 37.5 \\
     GPT 4o mini CoT & 42.3 & 39.9 & 38.0 & 44.4 & 36.8 & \textbf{36.1} & 39.6 \\
    GPT 4o mini DFS  & 46.0 & 43.8 & 44.3 & 41.1 & 34.9 & 34.4 & 40.8 \\
    GPT 4o CoT & 45.4 & 43.8 & 44.3 & \textbf{54.0} & \textbf{45.3} & 32.8 & 44.3 \\
    GPT 4o DFS & 46.6 & \textbf{53.6 }& \textbf{44.9} & 50.0 & 42.5 & 34.4 & \textbf{45.3} \\

    \bottomrule
    \end{tabular}
    \caption{FAC scores for different methods and conditions.}
    \label{tab:main_fac}
\end{table*}
\subsection{Evalution Metrics}
\noindent \textbf{Solvable Pass Rate.} We use the SoPR metric in StableToolBench in this project. We evaluate all instances three times and calculate the average scores and standard deviation. We use \texttt{gpt-4o} as the evaluator model.\footnote{Evaluation models used in ToolBench and StableToolBench are \texttt{gpt-3.5-turbo} and \texttt{gpt-4-turbo}, both legacy models now}.

\noindent \textbf{Final Answer Completeness.} Besides the Solvable Pass Rate scores in StableToolBench, we propose the new FAC score to measure whether the final answer completes the user query well. In tool learning, users ultimately need the final answer. How the tool-using model reaches that answer is less important. Therefore, it is more meaningful to measure whether the final answer matches the need of the user's request. 

To ensure the stability of the metric, we decide to use performant closed-sourced models for annotation and train an offline evaluator. We first randomly sample 3,714 samples from ToolBench training data and let \texttt{gpt-4o}, \texttt{gpt-4o-mini} and \texttt{gpt-4-turbo} to annotate whether the answer completes the user request, given only these two pieces of information. The model is required to answer either solved or unsolved. The annotation prompt is shown in \Cref{tab:prompt_fac}. We then sample 77 of these samples and invite two human annotators to annotate ground truth. Accuracy scores of these models are shown in \Cref{app:evaluation_results}. Finally, we train a \texttt{Llama-3.1-8B-Instruct} model using the C-RLFT method~\cite{wang2023openchat}. We use the \texttt{gpt-4-turbo} tag during inference because \texttt{gpt-4-turbo} performs best in the previous annotation process. Our model finally achieves 88.3\% accuracy against both annotators.

\subsection{Compare Real and Simulated Environments}\label{sec:compare_real_sim}
To compare the real and simulated environments, we first need to generate a set of queries. It is important to note that we cannot directly use the queries from the StableToolBench test set, as many of the ground-truth APIs are no longer available. In contrast, the simulated APIs are always accessible, which would introduce a bias between the two environments. To create new queries, we follow the approach used in ToolBench but without limiting the group or categories of APIs. Specifically, we sample 2-5 APIs from the entire set of available APIs and prompt \texttt{gpt-4o} to generate queries based on these APIs. The prompt is provided in Table~\ref{tab:prompt_query_generation}. We then filter out unsolvable queries following the StableToolBench methodology. Next, we run \texttt{gpt-4o} and \texttt{gpt-4o-mini} using the CoT and DFS methods on these queries in the real environment. We ask \texttt{gpt-4o} to assess whether all APIs are functioning correctly in the response trajectory, using the prompt shown in Table~\ref{tab:prompt_tool_call}. Any queries involving unavailable tools are filtered out. After this filtering process, 158 queries remain.

We run \texttt{gpt-4o}, \texttt{gpt-4o-mini}, and \texttt{\seqsplit{ToolLLaMA-v2}} using both CoT and DFS methods on these queries. We consider three environments: the real environment, the environment simulated by \texttt{gpt-4o}, and the environment simulated by our \modelname. The results are presented in \Cref{fig:compare_all}. As shown in the figures, our \modelname closely matches the real environment across all tool-using methods, while the performance of \texttt{gpt-4o} in simulated environments is significantly lower. This demonstrates that \modelname better simulates the tool environment compared to \texttt{gpt-4o}.

\subsection{StableToolBench Performance with \modelname-Cache}
As an improved version for StableToolBench environments, we evaluate baseline model performance with \modelname-Cache simulated environments.
Firstly, we evaluate the model on the cache data, which is also real calls in the test data of StableToolBench. We randomly sample 200 instances from the cache test set stated in \Cref{sec:method_cache}. Then we test the cache model on the test set. The BLEU-4 and observation following scores are $84.9$ and $8.82$ respectively, demonstrating a good performance on the cache API data. As a comparison, the general \modelname gets 27.1 and 4.76 respectively, showing the effectiveness of finetuning.

Then, we run \texttt{ToolLLaMA-v2}, \texttt{GPT-4o-mini}, and \texttt{GPT-4o} using both CoT and DFS inference methods on the simulated environments. The \texttt{single step max length} is set to 20, and the \texttt{max observation length} is set to 2048. The results are shown in \Cref{tab:main_sopr} and \Cref{tab:main_fac}. As observed in the tables, DFS methods generally outperform CoT methods, which aligns with the findings in ToolBench. Additionally, the latest \texttt{GPT-4o} model outperforms the other models, as expected. However, the performance gap between DFS and CoT methods is narrower than anticipated. This is likely due to the reduced number of failing APIs during the inference stages, making the retry mechanism in DFS less advantageous. Moreover, the FAC scores are higher than the pass rate scores, which is consistent with expectations. Despite this, even \texttt{GPT-4o DFS} achieves only 45.3, which remains far from satisfactory.




\section{Analysis}
\subsection{Ablation Studies}
\label{sec:ablation}
To evaluate the effectiveness of CoT training, we compare several models on the OOD Succ test set using the observation-following metric. The models tested include: our proposed \modelname, \modelname without CoT training, \modelname without augmented and CoT data, and a model trained sequentially with SFT and CoT. Note that no additional data was created specifically for CoT compared to SFT. For inference, we use the direct SFT prompt for all models except the one trained sequentially with CoT, which uses the CoT prompt to achieve better performance.

The results are presented in ~\Cref{tab:ablation_cot}. As shown, incorporating artificially augmented and CoT training data enhances model performance. Interestingly, generating outputs using the SFT prompt outperforms the CoT prompt, as detailed in \Cref{sec:results}. However, omitting CoT data during training negatively impacts performance. We hypothesize that adding CoT data lowers the training difficulty, as models may learn to reason implicitly during inference. This remains an open area for future research. Furthermore, the CoT training data is much smaller in size compared to the SFT data, which makes inference with CoT more challenging. Lastly, while sequential training with SFT and CoT yields worse performance than mixed training, it still outperforms training without CoT, supporting our hypothesis.

\subsection{Cache Model on General Tasks}
While \modelname-Cache achieves satisfactory results on the StableToolBench test set, it is important to note that this post-training process does not degrade performance on OOD successful tasks. Specifically, the observation following metrics for CoT and SFT are $6.13$ and $6.81$, respectively. These results suggest that the \modelname has learned a robust representation that generalizes well across the benchmark through the diverse training datasets used. In this context, post-training fine-tunes the model on more task-specific features, but these adjustments do not significantly alter the model's core capabilities, allowing it to maintain strong performance across a variety of tasks.

\subsection{Finetuning Reasoning Model}
It is intriguing to examine the performance of fine-tuned reasoning models, especially given that test-time scaling has shown remarkable results on complex tasks~\cite{deepseekai2025deepseekr1incentivizingreasoningcapability}. After fine-tuning the Deepseek-R1-Distill-Qwen model, we observed that its documentation and instruction following score on OOD successful tasks is significantly lower than that of the \modelname and the o1-preview, with scores of $5.54$ and $5.55$ for CoT and SFT, respectively. Two potential explanations for this underperformance are as follows. First, the distilled models are heavily tuned for reasoning tasks, which may limit their ability in new domains. Second, the reasoning required for generating API responses can be subtle or implicit, leading to inefficiencies. In contrast, the normal LLM can more effectively handle tasks without unnecessarily forcing a reasoning process, which may explain its better performance in these scenarios.

\begin{table}[t!]
    \centering
    \small
    \begin{tabular}{lc}
    \toprule
    \textbf{Method} & \textbf{Score} \\
    \midrule
        \modelname & 6.86 \\
        \quad w/ Seq CoT & 6.33 \\
        \quad w/o CoT & 6.70 \\
        \quad w/o A+CoT & 6.26 \\
        
    \bottomrule
    \end{tabular}
    \caption{Ablation on the CoT Training on the OOD Succ test set. A and CoT represent data augmentation and CoT training data respectively. Seq CoT means the model is trained on SFT data and then CoT data.}
    \label{tab:ablation_cot}
\end{table}
\section{Related Work}
\noindent\textbf{Tool Learning Benchmarks.} 
LLMs augmented with external tools have demonstrated remarkable problem-solving capabilities, often exceeding their stand-alone performance~\citep{li2023apibank, patil2023gorilla, gpt4tools, song2023restgpt, tang2023toolalpaca, ye2024tooleyes, xu2023tool}. 
Consequently, numerous benchmarks have been developed to evaluate proficiency in various aspects of tool learning, including tool selection~\citep{seal-tools, xu2023toolmanipulationcapabilityopensource}, task planning~\citep{huang2024planningcreationusagebenchmarking, shen2024taskbenchbenchmarkinglargelanguage}, API stability~\citep{guo-etal-2024-stabletoolbench} and query formulation~\citep{shen2025shortcutsbenchlargescalerealworldbenchmark}.

\noindent\textbf{Tool Environment Simulation.} 
Prior studies~\citep{lù2024weblinxrealworldwebsitenavigation, yao2023webshopscalablerealworldweb, rawles2024androidworlddynamicbenchmarkingenvironment} have explored various real-world tool environment. 
Recent work has shifted toward simulated environments, such as web simulators~\citep{pmlr-v70-shi17a, yao2023webshopscalablerealworldweb, zhou2024webarenarealisticwebenvironment, furuta2024exposinglimitationslanguagemodel, chae2024webagentsworldmodels}, tool simulators for safety scenarios~\citep{ruan2024identifyingriskslmagents}, and multi-modal environment simulation~\citep{zheng2024agentstudiotoolkitbuildinggeneral}. 
However, existing approaches struggle to balance stability, scalability, and realness. To address this, we propose a novel framework for training specialized LLMs to simulate tool environments effectively.




\section{Conclusion}
In this work, we present \modelname, a novel framework that trains specialized LLMs to act as ``mirrors'' to tool environments by accurately replicating real API responses. The simulator is trained on request-response data pairs from over 7,000 APIs sourced from RapidAPI, with performance enhanced through a novel API mechanism reasoning framework. Experimental results demonstrate that \modelname achieves strong documentation and instruction following capability. Responses provided by it are also the most similar to the real environment compared to baseline prompting methods. Furthermore, when deployed as an environment model in StableToolBench, \modelname delivers reliable outcomes comparable to those obtained from real-world APIs, validating its practical utility.

Beyond the immediate scope of this work, general tool simulators may be beneficial to advancing LLMs' tool-using capabilities. First, such simulators can enable execution feedback during training or inference without costly real-environment interactions, paving the way for novel online training algorithms and adaptive inference strategies~\cite{wang-etal-2024-llms-imaginarium, renze-2024-effect, yu2025exactteachingaiagents}. Second, they facilitate the generation of diverse training trajectories through the integration of synthetic APIs, addressing data scarcity challenges in tool-learning scenarios. Finally, while our current focus lies in simulating functional and reliable tools, the framework can be extended to simulate tools with operational failures, safety-critical vulnerabilities, or ethical concerns. This capability could significantly expand the scope of tool-learning research by enabling risk-aware training and robustness testing in controlled environments.

\section*{Limitations}

In this work, we propose a trained tool simulator specialized in mirroring real tool environments. However, our approach has certain limitations.
Firstly, real API requests often fail in practical environments due to issues such as connectivity problems. While it is crucial for a tool-using model to effectively handle such failures, this aspect is not addressed in our project. Nevertheless, our system can be easily adapted to simulate these scenarios by modifying the system prompts or intentionally introducing failures manually.
Secondly, as a general simulator, \modelname has the potential to provide immediate feedback to tool-using models, thereby enhancing their performance. This capability could also be leveraged to evaluate the effectiveness of the models in greater detail. However, due to time and resource constraints, we have not conducted experiments in this direction, leaving it as an avenue for future work.

\bibliography{custom,anthology}
\clearpage

\appendix

\section{Request writing}\label{app:request_writing}
An example of request writing is shown in \Cref{tab:example_request_writing}
\begin{table}[ht!]
    \resizebox{\columnwidth}{!}{
    \begin{tabular}{p{\linewidth}}
    \toprule
    \rowcolor[gray]{0.95}\multicolumn{1}{c}{\textbf{API Documentation}} \\
    \midrule
   \makecell[{{p{\linewidth}}}]{
    \textbf{API Name:} List Languages \\
    \textbf{API Desc.:} Get a list of currently supported languages. \\
    \textbf{Required Param:} N/A \\
    \textbf{Optional Param:} N/A \\
    \textbf{Tool Desc.:} This is a tool used to ... \\
    \textbf{Tool Name:} TTSKraken \\
    \textbf{Tool Category:} Artificial Intelligence
    } \\
    \midrule
    \rowcolor[gray]{0.95}\multicolumn{1}{c}{\textbf{Scenario}} \\
    \midrule
    \makecell[{{p{\linewidth}}}]{
     You are a content creator tasked with producing an engaging educational video for young learners that helps them understand basic math concepts in multiple languages. To make the video more accessible, you want to incorporate voiceovers in various languages. First, you need to gather a list of the languages supported by the TTSKraken API. Once you have the language list, you will choose three different languages that can target international audiences. Next, you will convert your script, which explains addition and subtraction in a fun way, into audio using the TTSKraken service in those selected languages. Finally, you will integrate these audio files into your video, aiming to enhance the learning experience for children from different linguistic backgrounds.,
 \\
    } \\
    \midrule
    \rowcolor[gray]{0.95}\multicolumn{1}{c}{\textbf{LLM-written request}} \\
    \midrule
    \makecell[{{p{\linewidth}}}]{
     \texttt{``category'': ``Artificial\_Intelligence''}, \\
    \texttt{``tool\_name'': ``TTSKraken''}, \\
    \texttt{``api\_name'': ``List Languages''}, \\
    \texttt{``tool\_input'': ``\{\}''}
} \\
\bottomrule

    \end{tabular}}
    \caption{An example of API documentation, scenario and LLM-written request.}
    \label{tab:example_request_writing}
\end{table}

\section{Distribution of Training Data}\label{app:distribution}
\begin{figure*}
    \centering
    \includegraphics[width=\linewidth]{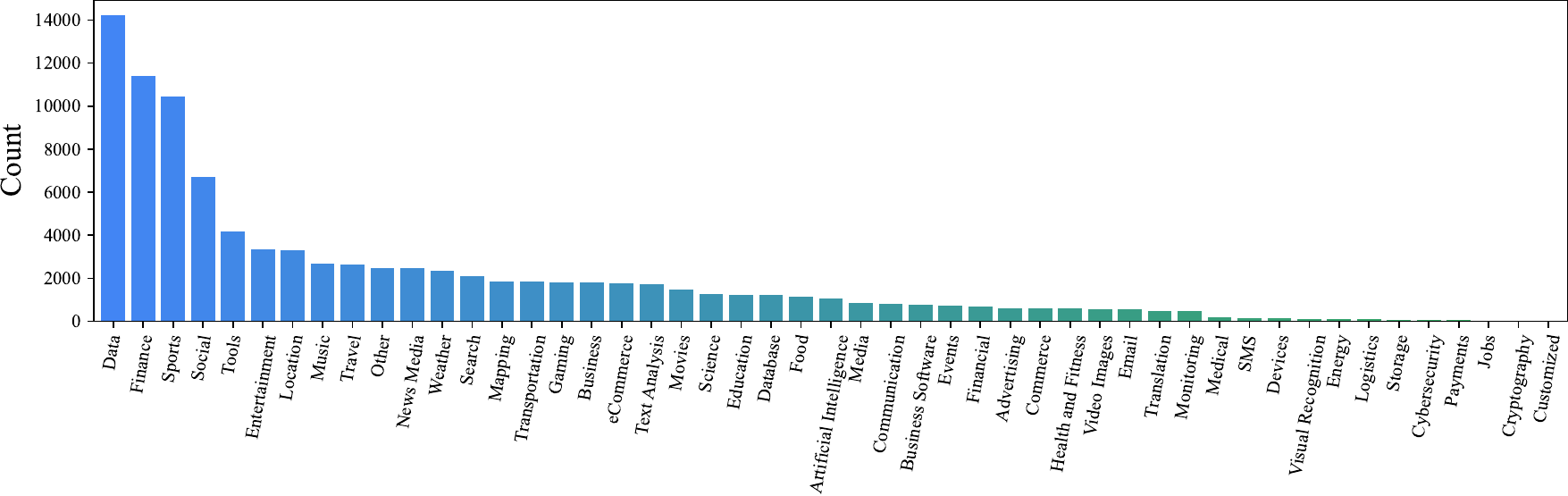}
    \caption{Data count in each tool category}
    \label{fig:tool_category_count}
\end{figure*}

The distribution of the training set is shown in \Cref{fig:tool_category_count}.

\section{Prompts}

\begin{table*}[htbp!]
    \centering
    \begin{tabular}{p{0.1\textwidth}p{0.8\textwidth}}
    \toprule
    \rowcolor[gray]{0.95} 
    \multicolumn{2}{c}{\textbf{Prompt for Checking Documentation Following}} \\
    \midrule
    System & \makecell[{{p{.8\textwidth}}}]{
We have a bunch of APIs at hand but their documentations may be out of date. Please help us evaluate the quality of an API response by assessing its adherence to API documentation.\\
You will be given the API documentation, containing the API name, description, tool name, tool description, required parameters, and optional parameters. (A tool may contain several APIs and has an overall functionality) You will also receive a user request (in a JSON format) and the API response (in the format of {"error": "error message if there is any, can be empty", "response": "the response content. may be empty if there is an error."}). \\
\# Goals: \\
Your primary goal is to determine how well the given API response adheres to the provided API documentation. \\
\# Notes: \\
- If the user request is malformed, a correct response may include an appropriate error message. This is to say if a user request is malformed and the response points it out, then the response is a good one. \\
- You need to only judge the API response, not the user request. \\
- As long as the response is a possible valid response following the documentation, it should be considered correct. \\
- Your analysis should remain neutral with an emphasis on objective evaluation. \\
\# Output: \\
Provide a detailed evaluation that reflects adherence to the above criteria, clearly identifying any deviations from the expected results. \\
Respond in the following JSON format: \\
\texttt{\{``overall\_eval'':1/0} (1 means meet the goal and 0 otherwise), \texttt{``reason'':``xxx''\}} \\
    } \\
    \hline
    User & \makecell[{{p{.85\linewidth}}}]{
\# API Documentation: \\
API Name: \texttt{\{api\_name\}} \\
API Description: \texttt{\{api\_description\}} \\
Tool (the API belongs to) Name: \texttt{\{tool\_name\}} \\
Tool (the API belongs to) Description: \texttt{\{tool\_description\}} \\
API Required Parameters: \texttt{\{required\_parameters\}} \\
API Optional Parameters: \texttt{\{optional\_parameters\}} \\
\# User Request: \\
\texttt{\{user\_request\}} \\
\# API Response: \\
\texttt{\{api\_response\}} \\
    } \\
    \bottomrule
    \end{tabular}
    \caption{Prompt for checking documentation following.}
    \label{tab:prompt_check_document_follow}
\end{table*}

\begin{table*}[htbp!]
    \centering
    \begin{tabular}{p{0.1\textwidth}p{0.8\textwidth}}
    \toprule
    \rowcolor[gray]{0.95} 
    \multicolumn{2}{c}{\textbf{API Simulation Prompt}} \\
    \midrule
    User & \makecell[{{p{.8\textwidth}}}]{
I have an API and call this API with a specific request. The API returned a response. \\
Can you give some hints about the mechanism behind the API, i.e. how the API works? This will be used to help an API simulator that behaves like the API you are reasoning about. Note that the simulator will not be able to see the real response so your hint should be instructive and constructive but not leak any specific information about the real response. \\
You will be given the API's documentation, parameters specification and the response returned by the API. \\
Make sure to limit your reasoning in 300 words. Return only your reasoning of how the API works, not the response. \\
API Name: \texttt{\{api\_name\}} \\
API Description: \texttt{\{api\_description\}} \\
Required Parameters: \texttt{\{required\_parameters\}} \\
Optional Parameters: \texttt{\{optional\_parameters\}} \\
Tool (The API belongs to) Name: \texttt{\{tool\_name\}} \\
Tool Description: \texttt{\{tool\_description\}} \\
API Request: \texttt{\{api\_request\}} \\
API Response: \texttt{\{api\_response\}} \\
    } \\
    \bottomrule
    \end{tabular}
    \caption{Rationale generation prompt.}
    \label{tab:prompt_rationale_generation}
\end{table*}

\begin{table*}[htbp!]
    \centering
    \begin{tabular}{p{0.1\textwidth}p{0.8\textwidth}}
    \toprule
    \rowcolor[gray]{0.95} 
    \multicolumn{2}{c}{\textbf{SFT Mode System Prompt}} \\
    \midrule
    System & \makecell[{{p{.8\textwidth}}}]{
    Imagine you are an API Server operating within a specialized tool, which contains a collection of distinct APIs. Your role is to deeply understand the function of each API based on their descriptions in the API documentation. As you receive specific inputs for individual API calls within this tool, analyze these inputs to determine their intended purpose. Your task is to craft a JSON formatted response that aligns with the expected output of the API. The JSON scheme is: \\
    \texttt{\{
        ``error'': ``'',
        ``response'': ``''
    \}}\\
The error field should remain empty, indicating no errors in processing. The response field should contain the content you formulate based on the API's functionality and the input provided. Ensure that your responses are meaningful, directly addressing the API's intended functionality. \\
The key is to maintain the JSON format's integrity while ensuring that your response is an accurate reflection of the API's intended output within the tool. \\
Please note that your answer should not contain anything other than a json format object, which should be parsable directly to json. \\
Note that: \\
- your response should contain rich information given the api input parameters. \\
- your response must be effective and have practical content. \\
API calls may fail for various reasons, such as invalid input parameters, authentication issues, or server errors. Your goal is to generate a response that accurately reflects the API's intended functionality, even if the input parameters are incorrect. Your response should be informative and relevant to the API's purpose, providing a clear and concise explanation of the expected output based on the input provided. \\
Here is an example: \\
API doc: \texttt{Example API Documentation} \\
Request: \texttt{Example Request} \\
Response: \texttt{Example Response} \\
    } \\
    \hline
    User & \makecell[{{p{.85\linewidth}}}]{
    API doc: \\
    \texttt{\{api\_doc\}} \\
    Request: \\
    \texttt{\{request\}} \\
    } \\
    \bottomrule
    \end{tabular}
    \caption{Supervised fine-tuning system prompt.}
    \label{tab:prompt_system_sft}
\end{table*}

\begin{table*}[htbp!]
    \centering
    \begin{tabular}{p{0.1\textwidth}p{0.8\textwidth}}
    \toprule
    \rowcolor[gray]{0.95} 
    \multicolumn{2}{c}{\textbf{CoT Mode System Prompt}} \\
    \midrule
    System & \makecell[{{p{.8\textwidth}}}]{
    [CHAIN\_OF\_THOUGHT] \\
You are an API Server operating within a specialized tool, tasked with understanding the purpose of each API based on provided documentation. Your job is to process specific API inputs and craft a well-formatted response reflecting the API's intended functionality. You should first infer the mechanism behind the API and then provide your response based on the input parameters. \\
Your response must follow this JSON structure: \\
\texttt{\{
    ``mechanism\_of\_the\_api'': ``'',
    ``error'': ``'',
    ``response'': ``''
\}}\\
* MECHANISIM OF THE API: Try to infer how the API functions based on the input parameters. \\
* ERROR: Leave empty unless there's an issue with the input. \\
* RESPONSE: Provide content based on the API's function. If examples are ineffective, give an independent, meaningful response.\\
Note: \\
* Ensure responses are practical, clear, and relevant. \\
* Handle incorrect input gracefully by explaining expected behavior. \\
Here is an example: \\
API doc: \texttt{Example API Documentation} \\
Request: \texttt{Example Request} \\
Response: \texttt{Example Response} \\
    } \\
    \hline
    User & \makecell[{{p{.85\linewidth}}}]{
    API doc: \\
    \texttt{\{api\_doc\}} \\
    Request: \\
    \texttt{\{request\}} \\
    } \\
    \bottomrule
    \end{tabular}
    \caption{Chain-of-Thought system prompt.}
    \label{tab:prompt_system_cot}
\end{table*}

\begin{table*}[htbp!]
    \centering
    \begin{tabular}{p{0.1\textwidth}p{0.8\textwidth}}
    \toprule
    \rowcolor[gray]{0.95} 
    \multicolumn{2}{c}{\textbf{Prompt Used In LLM-as-a-Judge }} \\
    \midrule
    System & \makecell[{{p{.8\textwidth}}}]{
You are a helpful assistant.
    } \\
    \hline
    User & \makecell[{{p{\linewidth}}}]{
{[Instruction]} \\
Act as an impartial judge to evaluate the quality of an AI API simulation output based on the provided API documentation and user request. 
Assess the simulation's accuracy in adhering to the documentation and fulfilling the user request. You will receive both a reference answer, representing a real API response, and the simulator's answer. 
The simulator's response does not need to match the reference answer exactly but must be faithful to the documentation and user request. 
The reference answer is just one possible output, but not the only one (You should not judge how similar the simulator's output is to the real one). The most important factor is whether the response is consistent with the API documentation and the user request. 
Pay attention to both the structure and the content of the response. Note that the response does not need to include all (even key) information in the documentation and the user request. 
As long as it is a reasonable response from the API, it should be rated as $10$. Begin your evaluation by comparing the simulator's response with the documentation and user request. Identify and correct any mistakes. Be as objective as possible. 
After providing your explanation, you must rate the response on a scale of $0$ to $10$ by strictly following this format: ``{[{[rating]}]}'', for example: ``Rating: {[{[$5$]}]}''. \\
{[Question]} \\
\texttt{\{question\}} \\
{[The Start of Reference Answer]} \\
\texttt{\{ref\_answer\_1\}} \\
{[The End of Reference Answer]} \\
{[The Start of Assistant's Answer]} \\
\texttt{\{answer\}} \\
{[The End of Assistant's Answer]}
    } \\
    \bottomrule
    \end{tabular}
    \caption{LLM Judge Prompt in documentation and instruction following.}
    \label{tab:prompt_llm_judge}
\end{table*}

\begin{table*}[htbp!]
    \centering
    \begin{tabular}{p{0.1\textwidth}p{0.8\textwidth}}
    \toprule
    \rowcolor[gray]{0.95} 
    \multicolumn{2}{c}{\textbf{FAC Scoring Prompt}} \\
    \midrule
    User & \makecell[{{p{.8\textwidth}}}]{
Given a query and an answer provided by an AI agent, you now need to determine the answer\_status of whether the well solved the query, i.e. whether the need of the query is satisfied. You need to output ``Unsolved'' or ``Solved'' and your reason. You must obey the following rules: \\
You should response ``Solved'' when: \\
1. If the answer well provides the information needed by the query, then it is ``Solved''. The answer does not need to be perfect, and it only needs to make a genuine attempt to address the query. \\
2. Consider only Completeness: \\
The answer attempts to address every part of the query, regardless of whether the information provided is factually correct or accurate, unless there is a severe factual error. \\
3. For Multi-part Queries: \\
For queries with multiple parts, all parts must be addressed for the answer to be considered ``Solved''. \\
4. Genuine Attempt : \\
The answer makes a genuine attempt to provide the requested information or perform the requested task for all parts of the query. This includes scenarios where the answer concludes that ``nothing'' is a reasonable response (e.g., when the requested information does not exist or is not available, or a possible answer of the query is nothing and the model answers nothing after reasonable attempts).  \\
You should response ``Unsolved'' when: \\
1. Refusal, Apology, or Non-engagement: \\
The answer includes a refusal or apology (e.g., ``I'm sorry, I can't help with that''). The answer does not directly engage with or address the query in any way. \\
2. Multi-part Queries: \\
If the query has multiple parts and at least one part is not well addressed. \\
3. Severe Factual Error: \\
If the answer contains a severe factual error that significantly impacts the usefulness of the information provided. \\
Additional Guidelines: \\
1. VERY IMPORTANT: DO NOT BE TOO HARSH. The model does not need to be perfect, and the answer does not need to be flawless. It only needs to make a genuine attempt to address the query. \\
2. DO NOT evaluate factual accuracy or correctness of the information provided based on your knowledge. Assume that the information provided is accurate and focus solely on whether the answer attempts to address all parts of the query, unless there is a severe factual error that conflicts common knowledge. \\
3. Focus on Final Answer: Only the final answer is provided and should be considered, disregarding any processes that were used to generate the answer. You only need to judge whether the information need is satisfied. \\
4. Answer Completion: The agent does not need to detail how it arrived at the answer, only that the answer itself is complete and attempts to address the query. \\
Here are some examples: xxxx \\
Now give your reason and answer status in the following format: \\
Answer Status xxx (can only be ``Solved'' or ``Unsolved'') \\
Reason \\
xxx
    } \\
    \bottomrule
    \end{tabular}
    \caption{Prompt used in the FAC score.}
    \label{tab:prompt_fac}
\end{table*}

\begin{table*}[htbp!]
    \centering
    \begin{tabular}{p{0.1\textwidth}p{0.8\textwidth}}
    \toprule
    \rowcolor[gray]{0.95} 
    \multicolumn{2}{c}{\textbf{Query Generation Prompt}} \\
    \midrule

    User & \makecell[{{p{\linewidth}}}]{
You will be provided with several tools, tool descriptions, all of each tool's available API functions, the descriptions of these API functions, and the parameters required for each API function. Your task involves creating a varied, innovative, and detailed user query that employ API functions of multiple tools. For instance, given three tools `nba\_news' `cat-facts' `hotels': `nba\_news' has API functions ``Get individual NBA source news'' and ``Get all NBA news'', `cat-facts' has API functions ``Get all facts about cat'' and ``Get a random fact about cats'', `hotels' has API functions ``properties/get-details (Deprecated)'', ``properties/list (Deprecated)'' and ``locations/v3/search''. Your query should articulate something akin to: ``I want to name my newborn cat after Kobe and host a party to celebrate its birth. Get me some cat facts and nba news to gather inspirations for the cat name. Also, find a proper hotel around my house in Houston Downtown for the party.'' This query exemplifies how to utilize API calls of all the given tools. A query that uses API calls of only one tool will not be accepted. Additionally, you must incorporate the input parameters required for each API call. To achieve this, generate random information for required parameters such as IP address, location, coordinates, etc. For instance, don't merely say `an address', but provide the exact road and district names. Don't just mention `a product', but specify wearables, milk, a blue blanket, a pan, etc. Don't refer to `my company', but invent a company name instead. The first seven of the ten queries should be very specific. \\
The query should combine API calls of different tools in various ways and include the necessary parameters. Note that you shouldn't ask `which API to use', rather, simply state your needs that can be addressed by these APIs. You should also avoid asking for the input parameters required by the API call, but instead directly provide the parameters in your query. \\
The final query should be complex and lengthy, describing a complicated scenario where all the provided API calls can be utilized to provide assistance within a single query. \\
You should first think about possible related API combinations, then give your query. Related\_apis are apis that can be used for a given query; those related apis have to strictly come from the provided api names. For each query, there should be multiple related\_apis. \\
Deliver your response in this JSON format: \\
\texttt{\{
``query'':$\dots$,
``related\_apis'':[[<tool name>, <api name>], [<tool name>, <api name>], [<tool name>, <api name>]]
\}} \\
Examples: \\
\texttt{\{Example\}}\\
These are only examples to show you how to write the query. Do not use apis listed in the above examples, but rather, use the ones listed below in the INPUT. \\
INPUT: \\
\texttt{\{tools\_json\}} \\
OUTPUT:
    } \\
    \bottomrule
    \end{tabular}
    \caption{Prompt for generating queries.}
    \label{tab:prompt_query_generation}
\end{table*}

\begin{table*}[htbp!]
    \centering
    \begin{tabular}{p{0.1\textwidth}p{0.8\textwidth}}
    \toprule
    \rowcolor[gray]{0.95} 
    \multicolumn{2}{c}{\textbf{Calls Writing Prompt}} \\
    \midrule

    User & \makecell[{{p{\linewidth}}}]{
I have an API at my hand and I want to use this API to solve a problem. Can you write a call to test if the API is reachable? Remember to include all the required parameters. You must follow the API documentation to write your call. Do not make up any parameters Please also note that a tool is a collection of APIs that are used to solve a specific problem. \\
Your answer should  be in  the following json format: \\
\texttt{\{\{} \\
\texttt{    ``category'':``'',} \\
\texttt{    ``tool\_name'':``'',} \\
\texttt{    ``api\_name'':``'',} \\
\texttt{    ``tool\_input'':`\{\{\}\}',} \\
\texttt{    ``strip'':``filter'',} \\
\texttt{\}\}} \\
Tool Documentation: \{document\}
    } \\
    \bottomrule
    \end{tabular}
    \caption{Prompt for writing tool calls.}
    \label{tab:prompt_tool_call}
\end{table*}



\section{Request Error Identification and Request Filtering Rule}\label{app:filter_rule}
In this study, we classify request errors and filter out invalid requests to RapidAPI based on specific keyword occurrences within the error or response messages. The errors are categorized as follows:
\begin{itemize}
    \item Not Connected Error:  This error is identified when the error message includes terms such as \texttt{HTTP}, or when the response contains phrases like \texttt{HTTP error}, \texttt{connection}, \texttt{rate limit} or \texttt{timeout}.
    \item Not Found Error: This occurs when the error or response message includes terms such as \texttt{not found}, \texttt{not available}, \texttt{API doesn't exists}, \texttt{Service Not Found}, \texttt{internal error} or a 404 error message;
    \item Parameter Change: This category is triggered when the error message or response refers to issues with parameters, parsing, or undefined variables.
    \item Parsing Error: This error is detected when the error message begins with the phrase \texttt{Function executing from}.
    \item Not Authorised: Errors falling under this category are identified by terms such as \texttt{authorize},  \texttt{unauthorized}, \texttt{blocked user}, \texttt{unsubscribe}, \texttt{credential}, \texttt{disabled for your subscription}, or any mention of 401 or 403 error codes.
    \item Other Errors: This category includes errors that contain any non-empty error message not falling into the aforementioned categories.
    \item Success: All other requests that do not result in errors are classified as successful.
\end{itemize}

\section{Training Hyperparameters}
We trained the \modelname with the following hyperparameter settings: a learning rate of 2e-05, a train batch size per device of 2. The training process used a seed value of 42, with a multi-GPU distributed setup across 8 devices, each equipped with 40GB A100 GPUs. Gradient accumulation was applied with 8 steps, leading to a total effective train batch size of 128. The optimizer used was Adam, with betas set to (0.9, 0.999) and epsilon set to 1e-08. For the learning rate schedule, a cosine annealing approach was employed, with a warmup ratio of 0.04 and 100 warmup steps. The model was trained for a total of 5 epochs.

\section{Human Annotation on the FAC evaluator}
Results of human annotation and model performance of the FAC evaluator are shown in \Cref{tab:evaluation_results}. R1 and R2 stand for two annotators.
\label{app:evaluation_results}

\begin{table}[htbp!]
    \centering
    \small
    \begin{tabular}{lccccccc}
        \toprule
        \textbf{Method} & \textbf{R1 Acc} & \textbf{R2 Acc}\\
        \midrule
         R1 & -- & 92.2\\
         R2 & 92.2 & --\\
         \midrule
         GPT 4o mini & 79.4 & 82.3 \\
         GPT 4o & 88.2 & 88.2 \\
        GPT 4 Turbo & \textbf{97.2} & \textbf{91.2} \\
        \midrule
         \textbf{Ours} & \underline{88.3} & \underline{88.3} \\
         \bottomrule
    \end{tabular}
    \caption{Evaluation model results}
    \label{tab:evaluation_results}
\end{table}

\section{Model Performance on StableToolBench with \modelname-Cache}
\label{app:rep_perf}
Performance of the models that are used in ToolBench is shown in \Cref{tab:model_performance_cache_rep_sopr} and \Cref{tab:model_performance_cache_rep_fac}. We use the replication data released by ToolBench so these models are run in real environments but evaluated with \texttt{gpt-4o} now in the Pass Rate metric. Note that these scores are much lower than those reported by ToolBench, probably because \texttt{gpt-4o} is much more strict than \texttt{gpt-3.5-turbo}. But the relative ranking of models keeps the same. Please also note that there are a lot of failing tools in the replication making the performance lower than our simulated environments. An example is as follows in ToolLLaMA v2 CoT: 
\begin{itemize}
    \item Step 4: 
    \begin{itemize}
        \item Name: \texttt{cifications\_by\_custom\_id}
        \item Arguments: 
        \begin{lstlisting}[basicstyle=\ttfamily\footnotesize]
{
  "phonecustomid": 123456
}
        \end{lstlisting}
        \item Response:
        \begin{lstlisting}[language=HTML, breaklines=true, basicstyle=\ttfamily\footnotesize]
<!DOCTYPE html>
<html>
<head>
    <meta charset="UTF-8" />
    <title>An Error Occurred: Internal Server Error</title>
        <style>xxx</style>
        </head>
        <body>
        <div class="container">
        <h1>Oops! An Error Occurred</h1>
        <h2>The server returned a "500 Internal Server Error".</h2>
        <p>
        Something is broken. Please let us know what you were doing when this error occurred.
        We will fix it as soon as possible. Sorry for any inconvenience caused.
        </p>
        </div>
        </body>
</html>
        \end{lstlisting}
    \end{itemize}
\end{itemize}

\begin{table*}[htbp!]
    \centering
    \small
    \begin{tabular}{lccccccc}
        \toprule
        \textbf{Method} & \textbf{I1 Inst} & \textbf{I1 Cat} & \textbf{I1 Tool} & \textbf{I2 Cat} & \textbf{I2 Inst} & \textbf{I3 Inst} & \textbf{Average} \\
        \midrule
    GPT 4 0613 CoT (Rep) & 18.8{\tiny $\pm{0.3}$} & 24.6{\tiny $\pm{0.3}$} & 16.2{\tiny $\pm{1.6}$} & 13.7{\tiny $\pm{1.1}$} & 16.0{\tiny $\pm{1.5}$} & 2.2{\tiny $\pm{0.8}$} & 14.3{\tiny $\pm{1.0}$} \\
    GPT 4 0613 DFS (Rep)  & 18.4{\tiny $\pm{1.3}$} & 29.8{\tiny $\pm{1.3}$} & 25.7{\tiny $\pm{1.2}$} & 20.2{\tiny $\pm{1.3}$} & 16.4{\tiny $\pm{0.4}$} & 2.2{\tiny $\pm{0.8}$} & 18.5{\tiny $\pm{0.9}$} \\
    ToolLLaMA CoT (Rep) & 5.1{\tiny $\pm{1.0}$} & 16.6{\tiny $\pm{0.8}$} & 8.0{\tiny $\pm{1.3}$} & 9.7{\tiny $\pm{1.1}$} & 7.5{\tiny $\pm{0.8}$} & 3.3{\tiny $\pm{0.0}$} & 8.4{\tiny $\pm{0.9}$} \\
    ToolLLaMA DFS (Rep) & 7.4{\tiny $\pm{0.5}$} & 16.1{\tiny $\pm{0.6}$} & 12.4{\tiny $\pm{0.3}$} & 9.9{\tiny $\pm{0.8}$} & 9.1{\tiny $\pm{0.4}$} & 0{\tiny $\pm{0.0}$} & 8.2{\tiny $\pm{0.6}$} \\
    \bottomrule
    \end{tabular}
    \caption{Solvable pass rate scores.  ``Inst'' and ``Cat'' stand for the Instruction and Category subsets. ``Rep'' stands for the official replication data from ToolBench.}
    \label{tab:model_performance_cache_rep_sopr}
\end{table*}

\begin{table*}[htbp!]
    \centering
    \small
    \begin{tabular}{lccccccc}
        \toprule
        \textbf{Method} & \textbf{G1 Inst} & \textbf{G1 Cat} & \textbf{G1 Tool} & \textbf{G2 Cat} & \textbf{G2 Inst} & \textbf{G3 Inst} & \textbf{Average} \\
        \midrule
    GPT 4 0613 CoT (Rep) & 25.2 & 30.1 & 22.2 & 25.8 & 26.4 & 3.3 & 22.1 \\
     GPT 4 0613 DFS (Rep) & 27.6 & 35.3 & 32.3 & 32.3 & 26.4 & 4.9 & 26.5 \\
   ToolLLaMA CoT (Rep) & 0.0 & 19.0 & 11.4 & 15.3 & 8.5 & 1.6 & 9.3 \\
    ToolLLaMA DFS (Rep) & 0.0 & 19.0 & 13.3 & 19.4 & 13.2 & 0.0 & 10.8 \\
    \bottomrule
    \end{tabular}
    \caption{FAC scores for replication models. ``Inst'' and ``Cat'' stand for the Instruction and Category subsets. }
    \label{tab:model_performance_cache_rep_fac}
\end{table*}

\section{AI Usage}
We acknowledge that we use LLMs to help with polishing words and tables in the papers.

\end{document}